\pdfoutput=1

\documentclass[11pt]{article}

\usepackage[final]{acl}

\usepackage{times}
\usepackage{latexsym}

\usepackage[T1]{fontenc}

\usepackage[utf8]{inputenc}

\usepackage{microtype}

\usepackage{inconsolata}

\usepackage{amsmath}
\usepackage{amssymb}
\usepackage{mathtools}
\usepackage{amsthm}

\usepackage[capitalize,noabbrev]{cleveref}

\usepackage{graphicx}
\usepackage{amsfonts,amsthm}
\usepackage{times}
\usepackage{latexsym}
\usepackage{comment}
\usepackage{graphicx}
\usepackage{xcolor}
\usepackage{color,soul}
\usepackage{booktabs}
\usepackage{mathtools}
\usepackage{linguex}
\usepackage{subfigure}
\usepackage{array}
\usepackage{float}
\usepackage{bm}
\usepackage{arydshln}
\usepackage{tikz}
\usepackage{multirow}
\usepackage{pmboxdraw}
\usepackage{breakurl}
\usepackage{cleveref}
\usepackage{subcaption}
\usepackage{annotate-equations}
\usepackage{inconsolata}
\usepackage{lipsum}
\usepackage{multicol}

\usepackage{amsmath,amsfonts,bm}

\def\eqref#1{equation~\ref{#1}}

\def\1{\bm{1}}

\def\vx{{\bm{x}}}

\def\mW{{\bm{W}}}

\DeclareMathAlphabet{\mathsfit}{\encodingdefault}{\sfdefault}{m}{sl}
\SetMathAlphabet{\mathsfit}{bold}{\encodingdefault}{\sfdefault}{bx}{n}

\definecolor{darkgreen}{HTML}{138808}
\definecolor{green_drawio}{HTML}{82B366}
\definecolor{dark_green_drawio}{HTML}{557543}
\definecolor{dark_red_drawio}{HTML}{990000}
\definecolor{blue_drawio}{HTML}{6C8EBF}
\definecolor{orange_drawio}{HTML}{D79B00}
\definecolor{red_drawio}{HTML}{990000}
\definecolor{grey_drawio}{HTML}{303030}

\definecolor{blue_plotly}{HTML}{636EFA}
\definecolor{red_plotly}{HTML}{EF553B}

\title{On the Similarity of Circuits across Languages:\\
a Case Study on the Subject-verb Agreement Task}

\author{Javier Ferrando$^{1}$ \quad \ Marta R. Costa-jussà$^{2}$  \\
 $^{1}$Universitat Politècnica de Catalunya\\
 $^{2}$FAIR, Meta \\
  \texttt{jferrandomonsonis@gmail.com}}

\begin{document}
\maketitle
\begin{abstract}
\vspace{-5pt}
Several algorithms implemented by language models have recently been successfully reversed-engineered. However, these findings have been concentrated on specific tasks and models, leaving it unclear how \textit{universal} circuits are across different settings. In this paper, we study the circuits implemented by Gemma 2B for solving the subject-verb agreement task across two different languages, English and Spanish. We discover that both circuits are highly consistent, being mainly driven by a particular attention head writing a `subject number' signal to the last residual stream, which is read by a small set of neurons in the final MLPs. Notably, this subject number signal is represented as a direction in the residual stream space, and is language-independent. We demonstrate that this direction has a causal effect on the model predictions, effectively flipping the Spanish predicted verb number by intervening with the direction found in English. Finally, we present evidence of similar behavior in other models within the Gemma 1 and Gemma 2 families.\footnote{The code accompanying this work can be found at \\ \url{https://github.com/javiferran/circuits_languages}.}
\end{abstract}

\section{Introduction}
The widespread use of large language models (LLMs; \citealp{gpt3, hoffmann2022an, palm}) highlights the importance of research dedicated to interpreting how these models work internally~\citep{ferrando2024primer}, especially to ensure they are safe. Mechanistic interpretability (MI)~\citep{olah2022mechinterp} aims to reverse-engineer the algorithms implemented by language models. A large set of MI works have focused on circuit analysis~\citep{räuker2023transparent}, which locates subsets of components responsible for a behavior while giving human-understandable explanations of their roles. This research has made progress in identifying circuits that handle different tasks~\citep{wang2023interpretability,docstring,stolfo-etal-2023-mechanistic,stolfo2023understanding,geva-etal-2023-dissecting,hanna2023does}. However, it remains unclear whether the findings obtained through circuit analysis transfer to different settings. For instance, if different models learn similar circuits for solving the same task, or if models find different solutions for the same task in two different languages. In this work, we study the latter question. Via the subject-verb agreement (SVA) task~\citep{linzen-etal-2016-assessing,goldberg2019assessing}, we study the main components in Gemma models~\citep{gemmateam2024gemmaopenmodelsbased,gemmateam2024gemma} that are responsible across both English and Spanish.

\begin{figure*}[!t]
\begin{centering}\includegraphics[width=0.88\textwidth]{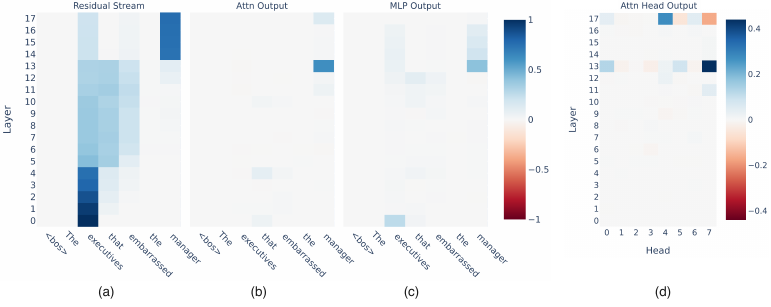}
\vspace{-8pt}
	\caption{English dataset activation patching results on the logit difference metric on (a) the residual streams (b) attention blocks outputs, (c) MLP outputs, and (d) on attention heads at the last position.}
	\label{fig:eng_merged}
	\end{centering}
\end{figure*}

\section{Experimental Setup}
In our experiments, we use several models of the Gemma 1 and 2 families~\citep{gemmateam2024gemmaopenmodelsbased,gemmateam2024gemma}. These models have a large vocabulary size (256k tokens), making it particularly well-suited for circuit analysis, especially when doing activation patching~(\Cref{sec:activation_patching}) in a multilingual setting, since it has a large set of non-English words with a reserved token. The analysis is focused on the Gemma 2B model, and we present evidence of similar behavior in a larger model (Gemma 7B) and in a model of the Gemma 2 family, Gemma 2 2B. Regarding the dataset, for the English experiments we use the subject-verb agreement (SVA) dataset from~\citet{arora2024causalgym}\footnote{\texttt{aryaman/causalgym}, subset \texttt{agr\_sv\_num\_subj-relc}}, built on top of SyntaxGym~\citep{gauthier-etal-2020-syntaxgym}. The dataset consists of contrastive pairs that differ in the subject number, which agrees with the verb form continuation. This allows us to create `clean' and `corrupted' versions:
\vspace{1pt}
\begin{equation}\label{ex:eng_contrastive_pairs}
\small
\begin{aligned}
&\text{\textit{Clean}: The} \eqnmarkbox[red_plotly]{subj_s}{\text{executive}} \text{that embarrassed the manager} \eqnmarkbox[red_plotly]{verb_s}{\text{\underline{has}}}\\
&\text{\textit{Corrupted}: The} \eqnmarkbox[blue_plotly]{subj_p}{\text{executives}} \text{that embarrassed the manager \rule{0.4cm}{0.1mm}}
\annotatetwo[yshift=0em]{above, label below}{subj_s}{verb_s}{Singular}
\annotate[yshift=0em]{below, label above}{subj_p}{Plural}
\end{aligned}
\end{equation}

\vspace{-7pt}


\section{Methods}\label{sec:methods}

We start searching for a circuit in Gemma 2B for solving the SVA in English. To do so we use common techniques in circuit analysis, mainly direct logit attribution, activation patching, and attention pattern analysis.

\paragraph{Direct Logit Attribution.} Every model component adds a vector $f^c(\mathbf{x})$ to the residual stream, and the last residual stream state gets projected onto the unembedding matrix, producing the logits distribution. Due to the linearity of the residual stream, the direct effect of a component to the logits can be measured by projecting its output onto the unembedding matrix, $f^c(\mathbf{x})\mW_{U}$. We can also measure the \textbf{d}irect \textbf{a}ttribution to the \textbf{l}ogit \textbf{d}ifference (DLDA)~\citep{yin-neubig-2022-interpreting,wang2023interpretability} of the two possible verb continuations ($g$ and $b$):
\begin{equation}\label{eq:dlda}
   \text{DLDA}_c = f^c(\tilde{\mathbf{x}})\mW_{U}[:,g] - f^c(\tilde{\mathbf{x}})\mW_{U}[:,b].
\end{equation}
\paragraph{Activation Patching.}\label{sec:activation_patching} A Transformer LM can be seen as a directed acyclic graph (DAG) representing a causal model~\citep{geiger2021causal,pearl_2009,causal_mediation_bias}, where nodes are model components, and edges representations. During the forward pass on the \textit{corrupted input} $\mathbf{x}$ we can intervene on the value of a node, $f^c(\mathbf{x})$, or residual stream state, $f^l(\mathbf{x})$ by taking the activation value from the forward pass on the \textit{clean input} $\tilde{\mathbf{x}}$. This is referred to as \textit{denoising activation patching}~\citep{causal_mediation_bias,meng2022locating}. We can express the intervention using the do-operator~\citep{pearl_2009} as $f(\mathbf{x}|\text{do}(f^c(\mathbf{x})=f^c(\tilde{\mathbf{x}})))$. Via a metric $m$ we measure how the prediction changes between both runs:
\begin{equation}\label{eq:patching_delta}
\text{AP}_c = m\bigl(f(\mathbf{x}), f(\mathbf{x}|\text{do}(f^c(\mathbf{x})=f^c(\tilde{\mathbf{x}})))\bigr).
\end{equation}
We are interested in finding components that increase the clean verb prediction when patching on the corrupted run. Thus, a natural choice for the patching metric $m$ is the logit difference between the clean and the corrupted verbs' logits. In the Example~\ref{ex:eng_contrastive_pairs}, this means computing the logit difference between ‘has’ and ‘have’, and we expect it to increase as we patch activations from the clean (which includes `executive') into the corrupted forward pass.

\begin{figure}[!t]
	\begin{centering}\includegraphics[width=0.49\textwidth]{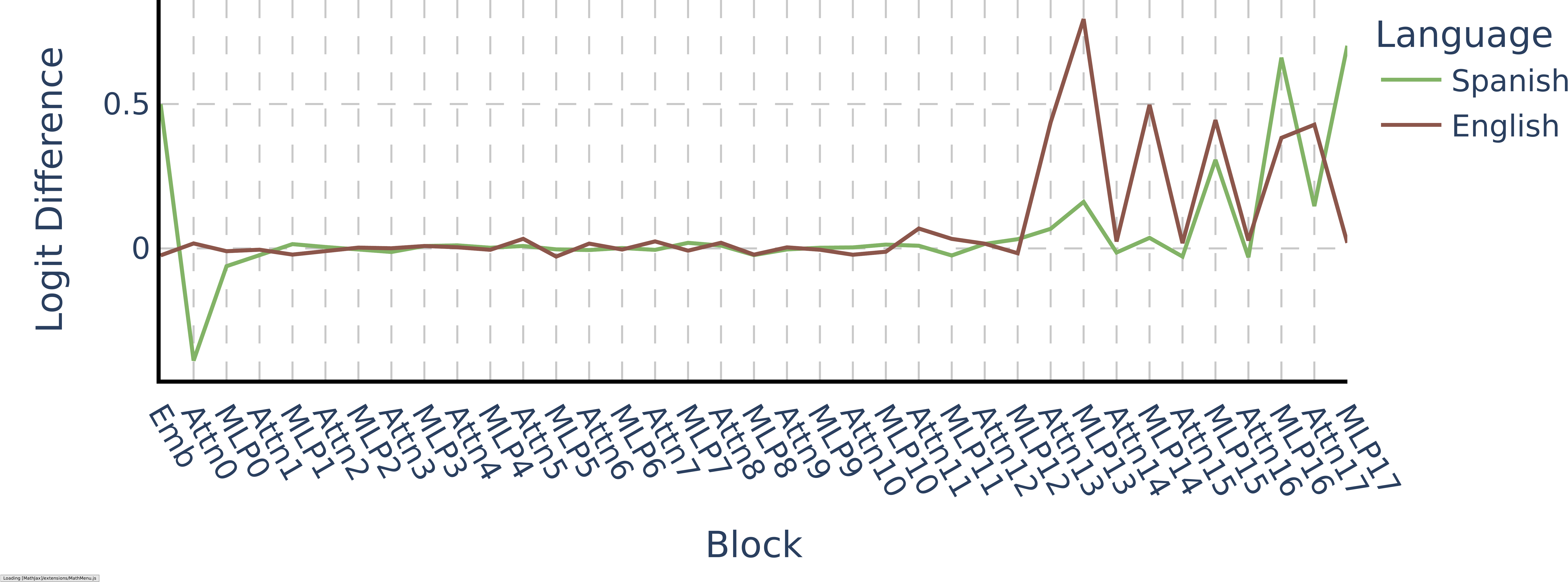}
	\caption{Average contribution to the logit difference by each model component.}
	\label{fig:both_components_contribs}
	\end{centering}
\end{figure}

\section{English Subject-Verb Agreement Circuit}\label{sec:english_sva_circuit}

\paragraph{Locating relevant components and residual stream states.} We perform activation patching on the residual stream states across the dataset and show the average logit differences\footnote{See in~\Cref{appx:logit_diffs} the average logit differences.} in~\Cref{fig:eng_merged} (a). We can see that the noun in the subject largely impacts the prediction, and patching at its position in early layers causes the verb prediction to aggressively change to match its number. Information from the subject flows towards the last residual stream via the attention block at layer 13~(\Cref{fig:eng_merged} (b)), followed by some action from downstream MLPs at the last position~(\Cref{fig:eng_merged} (b)), especially MLP at layer 13 (MLP13). We can also observe that ‘that’ and the following verb (`embarrassed') get information from the subject at middle layers. We get a more granular understanding of the attention layers that seem relevant by doing activation patching on the output of every attention head in the last position~(\Cref{fig:eng_merged} (d)). Attention head 7 in layer 13 (L13H7) has the largest effect on the logit difference, followed by L17H4. Notably, we also observe a head (L17H7) that contributes negatively to the logit difference. In~\Cref{appx:attention_patterns} we show the average output-value-weighted heatmaps of these heads, and we see that L13H7 attends broadly to the context, with a slight focus on `what', while L17H4 focuses on the subject's noun. Although attention blocks at layers 13 and 17 also have large direct effects~\Cref{fig:both_components_contribs}, most of the direct contribution to the logit difference is carried by downstream MLPs, specifically MLP14, MLP15, MLP16, and most notably MLP13.

\begin{figure}[!t]
	\begin{centering}\includegraphics[width=0.48\textwidth]{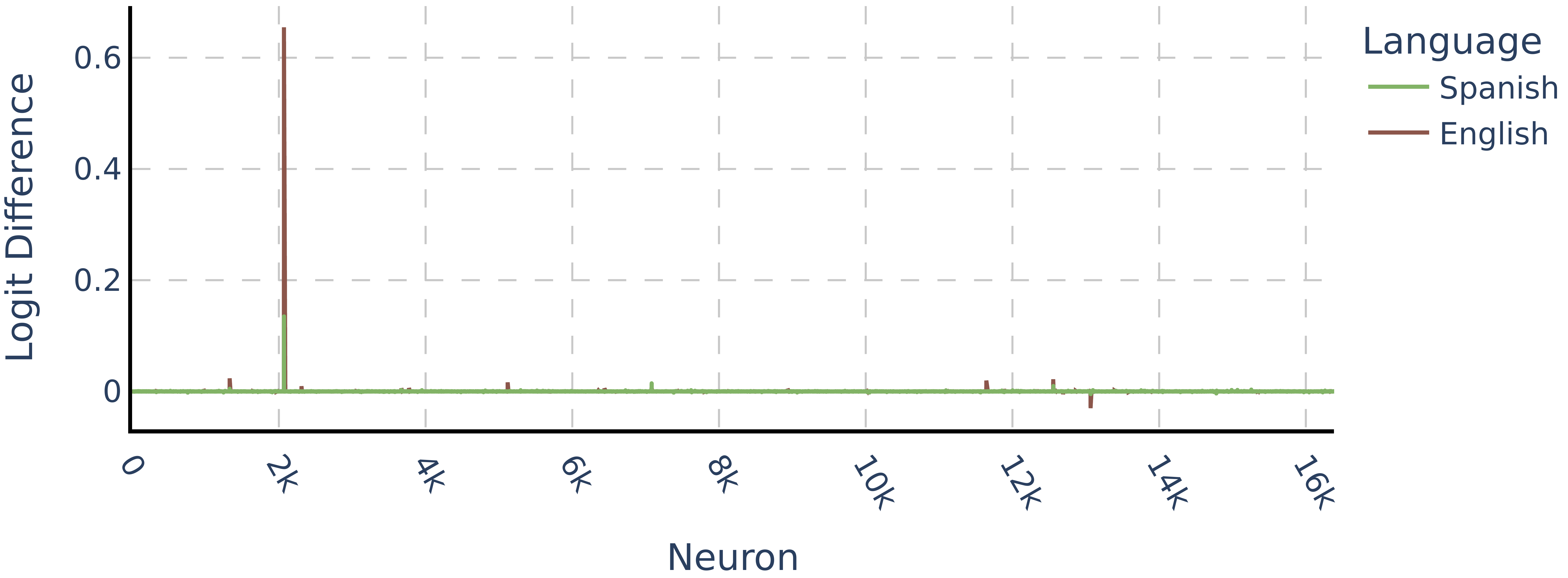}
	\caption{Average contribution to the logit difference by each neuron in MLP13.}
	\label{fig:mlp13_neurons}
	\end{centering}
\end{figure}

\begin{table}[!t]
    \centering
   \resizebox{0.42\textwidth}{!}{%
    \begin{tabular}{l} \toprule
    \multicolumn{1}{l}{\textbf{Top Promoted Tokens \textcolor{blue_plotly}{\textit{Positive}} Neuron Activation}}\\ 
\textbf{` are'}, \textbf{`are'}, \textbf{`were'}, \textbf{` were'}, \textbf{`Are'}, \textbf{`aren'}, \textbf{` ARE'},\\
\textbf{` WERE'}, \textbf{` weren'}\\
\midrule
\multicolumn{1}{l}{\textbf{Top Promoted Tokens \textcolor{red_plotly}{\textit{Negative}} Neuron Activation}}\\ 
` gardent', \textbf{` is'}, \textbf{` has'}, ` sembrano', \textbf{` was'}, ` continúan',\\
` appartienment', \textbf{` isn'}, \textbf{` hasn'}, ` sostu'\\
\bottomrule
    \end{tabular}}
    \caption{Top promoted tokens by neuron 2069 in MLP13 based on the sign of the neuron.}
    \label{table:promoted_tokens_n2069}
\end{table}
\paragraph{Analysis of Neurons.}
The contribution of MLP13 to the logit difference is led by a single neuron (2069)~(\Cref{fig:mlp13_neurons}). Recall that Gemma models use gated MLPs, which compute
 \begin{equation}\label{eq:gated_mlp}
    \text{GMLP}(\vx) = \bigl(\underbrace{g(\vx \mW_{\text{gate}} ) \odot \vx\mW_{\text{in}}}_{\text{neurons}} \bigr) \mW_{\text{out}},
\end{equation}
where $g$ is the activation function (GeGLU), $\mW_{\text{gate}}, \mW_{\text{in}} \in \mathbb{R}^{d \times d_{\text{mlp}}}$ read from the residual stream, and the linear combination of the rows of $\mW_{\text{out}} \in \mathbb{R}^{d_{\text{mlp}} \times d}$ weighted by the neuron values is added back to the residual stream~(see \Cref{apx:gmlp} for a visual description). This means that, unlike standard MLPs, neurons in GMLPs can take arbitrarily large positive and negative values. In the case of neuron 2069 in MLP13, when the neuron positively activates, their associated neuron weights (row in $\mW_{\text{out}}$) write in the direction of plural verb forms (and suppress singular forms)~(\Cref{table:promoted_tokens_n2069}). On the other hand, on negative activations, the neuron weights write in the direction of singular verb forms (and suppresses plural forms). Notably, this is true for the English and the Spanish verbs in our datasets, which are present and past tenses of the verbs `to be' and `have', but we also observe less common non-English plural verb forms promoted on negative neuron activations. This neuron seems to read a ‘subject number’ signal, but where does this signal come from? A candidate is L13H7, which has a large total effect on the logit difference.

\begin{figure}[!t]
	\begin{centering}\includegraphics[width=0.35\textwidth]{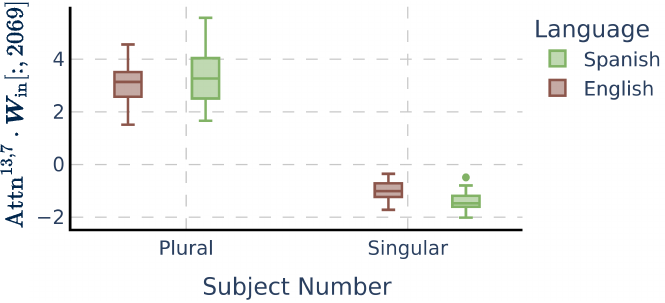}
	\caption{Dot product of the output of attention head L13H7 and the input weights of neuron 2069 in MLP13.}
 \label{fig:both_MLP13_neuron2069_act_subj_num}
	\end{centering}
\end{figure}

\begin{figure}[!t]
	\begin{centering}\includegraphics[width=0.48\textwidth]{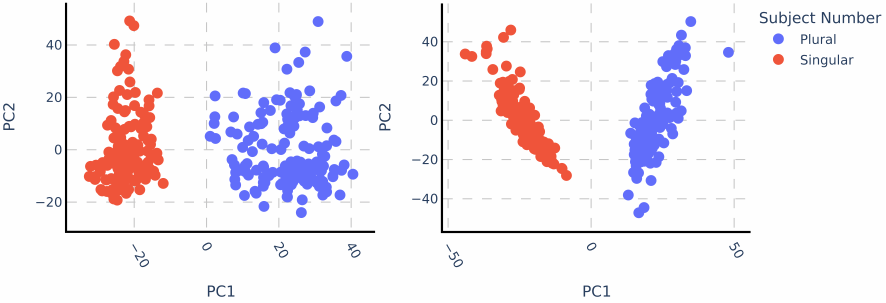}
	\caption{Projections of L13H7 outputs onto the top 2 PCs on English (left) and Spanish (right) dataset.}
	\label{fig:joint_pca_L13H7}
	\end{centering}
\end{figure}

\begin{figure*}[!t]
\begin{centering}\includegraphics[width=0.88\textwidth]{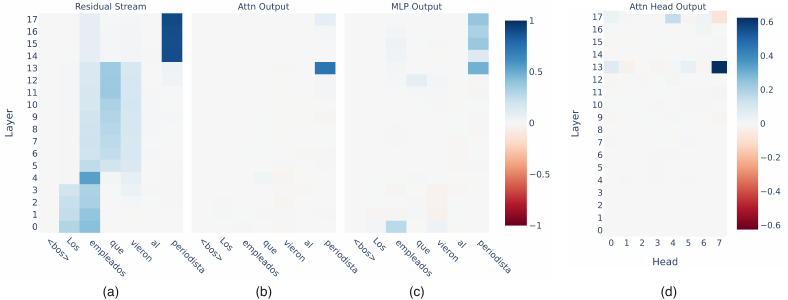}
\vspace{-8pt}
	\caption{Spanish dataset activation patching results on the logit difference metric on (a) the residual streams (b) attention blocks outputs, (c) MLP outputs, and (d) on attention heads at the last position.}
	\label{fig:spa_merged}
	\end{centering}
\end{figure*}

\paragraph{An Attention Head Writes the Subject Number in a Single Direction.} We compute the dot product between the output of attention head L13H7 at the last position and column 2069 of $\mW_{\text{in}}$ (${\mW_{\text{in}}[:,2069]}$) across the whole dataset and show the results in~\Cref{fig:both_MLP13_neuron2069_act_subj_num}. When the subject is singular, we get a negative dot product (activation) and promote singular verb forms~(\Cref{table:promoted_tokens_n2069}). When the subject is plural, we get positive dot product values and promote plural forms. We observe a similar pattern in other influential MLP neurons~(\Cref{sec:dot_products_L13H7}). We further provide evidence of the role of L13H7 by applying PCA on its outputs in the last residual stream~(\Cref{fig:joint_pca_L13H7}). The first principal component (PC1) clearly distinguishes between singular and plural subject examples. This means that L13H7 writes into a 1-dimensional subspace where the subject number signal is encoded, from which downstream neurons read to promote the correct tokens.

\section{Spanish Subject-Verb Agreement Circuit}

To study the subject-verb agreement task in Spanish, we follow the style of the English dataset, where we first prompt GPT4~\citep{openai2024gpt4} to generate verbs and nouns, and remove those words tokenized into multiple subwords. Then, we build similar examples to the ones in the English dataset. An example of a contrastive pair is:
\vspace{9pt}
\begin{equation}\label{ex:spa_contrastive_pairs}
\small
\begin{aligned}
&\text{\textit{Clean}: El} \eqnmarkbox[red_plotly]{subj_s}{\text{ingeniero}} \text{que ayudó al cantante} \eqnmarkbox[red_plotly]{verb_s}{\text{\underline{era}}}\\
&\text{\textit{Corrupted}: Los} \eqnmarkbox[blue_plotly]{subj_p}{\text{ingenieros}} \text{que ayudaron al cantante } \rule{0.4cm}{0.1mm}\\
\annotatetwo[yshift=0em]{above, label below}{subj_s}{verb_s}{Singular}
\annotate[yshift=0em]{below, label above}{subj_p}{Plural}
\end{aligned}
\end{equation}

\vspace{-8pt}
\paragraph{Spanish circuit is consistent with the English circuit.} With activation patching we see a similar pattern to that of the English dataset. Information from the subject flows to the last residual stream at layer 13, where the attention block shows a large effect~(\Cref{fig:spa_merged}). Also similarly, downstream MLPs are relevant for correctly solving the task, with MLP13 showing the highest total effect~(\Cref{fig:spa_merged} (b)), while MLP15, MLP16 and MLP17 having large direct effects on the logit difference. The contribution of MLP17 is notably greater than in the English dataset~(\Cref{fig:both_components_contribs}), where we observe non-English specific neurons~(\Cref{appx:neuron_1138_mlp17}). Activation patching on individual attention heads~(\Cref{fig:spa_merged} (d)) shows that, as in the English dataset, attention heads L13H7 and L17H4 have a positive influence on the correct verb form, while L17H7 influences negatively.
\begin{figure}[!t]
\begin{centering}\includegraphics[width=0.485\textwidth]{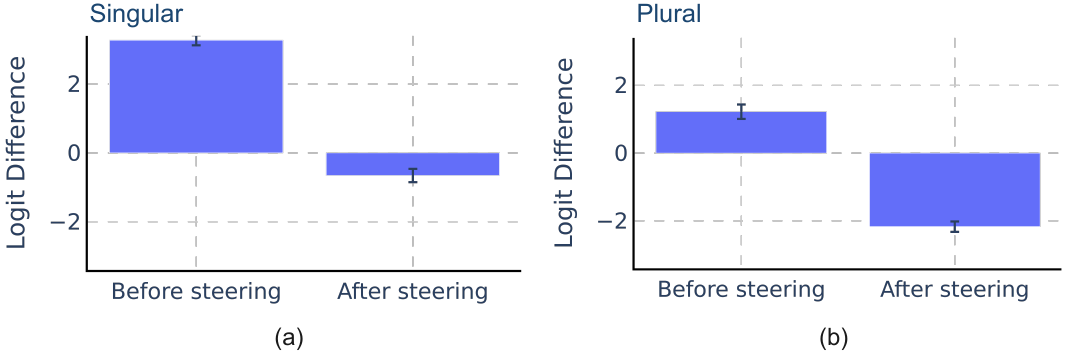}
	\caption{Spanish average logit difference in (a) singular subject and (b) plural subject examples, before and after steering the prediction with $\text{PC1}_{\text{English}}$.\vspace{-10pt}}
	\label{fig:spanish_sing_plur_logit_diffs_steered_pca}
	\end{centering}
\end{figure}

\paragraph{Activation Steering.} In both languages, the same attention head (L13H7) composes with specific neurons in downstream MLPs that are responsible for the correct verb form prediction, suggesting that this head writes a `subject number' signal, which is found via PC1~(\Cref{fig:joint_pca_L13H7}). Here, we study whether this direction, found in the English training set examples ($\text{PC1}_{\text{English}}$) has a causal effect on the model predictions, also on Spanish sentences. Specifically, we do activation steering~\citep{turner2023activation,li2023inferencetime,tigges2023linear} on the attention head output at the last position ($n$)
\begin{equation}
    \text{Attn}^{13,7}_n = \text{Attn}^{13,7}_n \pm \alpha \text{PC1}_{\text{English}},
\end{equation}
where $\alpha$ is a scalar coefficient controlling the magnitude of the steering effect on the unit norm $\text{PC1}_{\text{English}}$ vector. We select an appropriate $\alpha$ value in a validation set. Results show that adding $\text{PC1}_{\text{English}}$ successfully flips the Spanish verb number prediction to plural~(\Cref{fig:spanish_sing_plur_logit_diffs_steered_pca} (a)) on examples with singular subject, and that subtracting $\text{PC1}_{\text{English}}$ flips the Spanish plural number prediction to singular. Furthermore, we observe that the top predicted tokens other than verbs remain mostly unchanged~(see example in~\Cref{apx:predicted_tokens_steering}).

\begin{figure*}[!t]
\begin{centering}\includegraphics[width=0.98\textwidth]{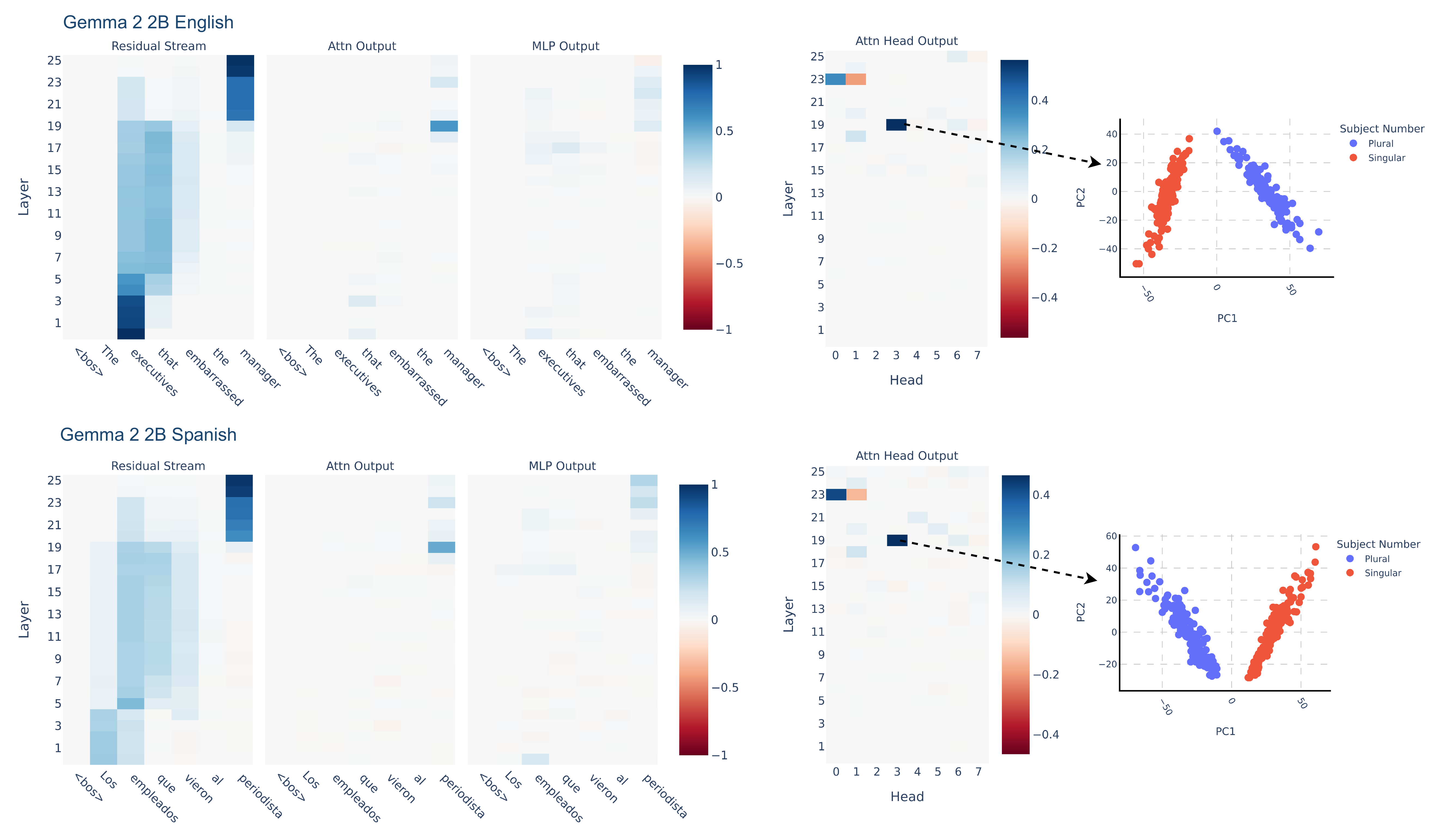}
\vspace{-8pt}
	\caption{Gemma 2 2B English (top) and Spanish (bottom) activation patching results. On the right it is shown the outputs of the most relevant attention head (L19H3) in PCA basis. We observe similar patterns to those in Gemma 2B, having a relevant attention head writing the subject number signal into a single direction (PC1).}
	\label{fig:gemma_2_2b_eng_spa_act_patching}
	\end{centering}
\end{figure*}

\section{Replicating Results in Gemma 7B and Gemma 2 2B}
To validate the generalizability of our findings across different model sizes and architectures, we extend our analysis to Gemma 7B and Gemma 2 2B. We perform activation patching experiments on both models, following the same methodology used for Gemma 2B.
Our results demonstrate large consistency in the circuit responsible for subject-verb agreement across these models. In both Gemma 7B and Gemma 2 2B, we observe similar patterns of information flow and the presence of a key attention head that plays a crucial role in encoding the subject number signal.

\begin{figure}[!t]
\begin{centering}\includegraphics[width=0.485\textwidth]{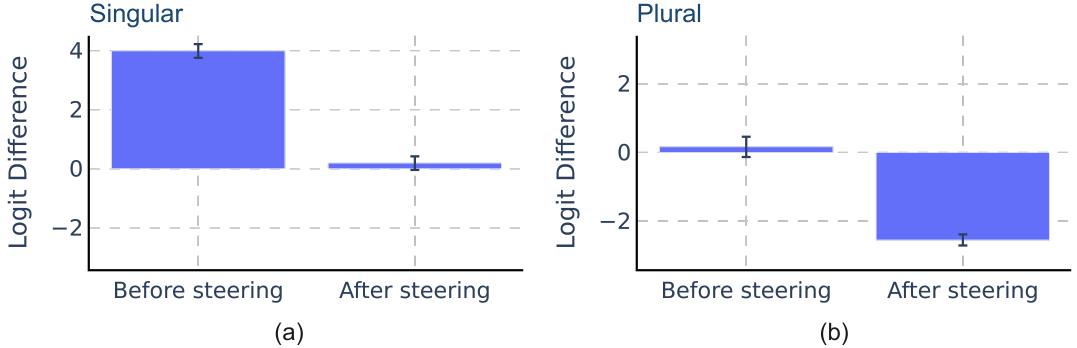}
	\caption{Gemma 2 2B Spanish average logit difference in (a) singular subject and (b) plural subject examples, before and after steering the prediction with $\text{PC1}_{\text{English}}$.\vspace{-10pt}}
	\label{fig:spanish_sing_plur_logit_diffs_steered_pca_gemma_2_2b}
	\end{centering}
\end{figure}

For Gemma 2 2B~(\Cref{fig:gemma_2_2b_eng_spa_act_patching}), we find that the attention head L19H3 serves a similar function to L13H7 in Gemma 2B. This head appears to be primarily responsible for writing the subject number signal into the residual stream. Notably, when we apply PCA to the outputs of this attention head, we discover that the subject number information is also encoded in the first principal component direction, just as we observed in Gemma 2B. Similarly, in Gemma 7B, we identify the attention head L22H6 as playing a analogous role~(\Cref{apx:results:gemma7b},~\Cref{fig:gemma_7b_act_patching_pca}). The activation patching results for both English and Spanish datasets in Gemma 2 2B closely mirror those of Gemma 2B, with the subject number signal being represented in the first principal component of the relevant attention head's output. Steering with $\text{PC1}_{\text{English}}$ also reduces significantly the logit difference in Spanish~(\Cref{fig:spanish_sing_plur_logit_diffs_steered_pca_gemma_2_2b}), even completely flipping the predictions in the plural case.

These findings suggest that the mechanism for handling subject-verb agreement is largely preserved across different scales and iterations of the Gemma model family. The consistent presence of a single attention head responsible for encoding subject number in a one-dimensional subspace indicates that this might be a fundamental architectural feature that emerges during training, regardless of model size or specific architecture details. This replication suggests that the circuit we've identified for subject-verb agreement may be a general feature of how these language models process grammatical number. Further research could explore whether this pattern holds true for even larger models or across other model families.

\section{Related Work}
Our work builds upon and extends several key studies in the field of probing~\citep{belinkov_probing} language models for syntactic knowledge, particularly focusing on agreement mechanisms in multilingual contexts.

Causal probing studies have provided evidence for the existence of specific syntactic agreement neurons in language models~\citep{finlayson-etal-2021-causal,de-cao-etal-2022-sparse,lakretz-etal-2019-emergence}. Furthermore, research has demonstrated that models like BERT~\citep{devlin-etal-2019-bert} rely on a linear encoding of grammatical number to solve the number agreement task~\citep{lasri-etal-2022-probing}. However, these studies have primarily focused on monolingual models, leaving a gap in our understanding of multilingual contexts.



Moving beyond monolingual studies, \citet{chi-etal-2020-finding} investigated universal grammatical relations in multilingual BERT. They developed a structural probe~\citep{hewitt-manning-2019-structural}, learning a linear mapping to a syntactic subspace where syntactic features overlap between languages. This study demonstrated the potential for identifying cross-lingual syntactic similarities in multilingual models.~\citet{mueller-etal-2022-causal} conducted a causal analysis of syntactic agreement neurons in multilingual language models. Via counterfactual interventions, they discovered significant overlap between languages in terms of neurons that causally influence syntactic agreement.

Recent research on mechanistic interpretability has shed light on how language models process multilingual inputs.~\citet{wendler-etal-2024-llamas} demonstrated that while processing non-English text, the English version of the next token can be decoded from middle layers. Building on this,~\citet{dumas2024how} provided evidence for language models developing universal concept representations that are disentangled from specific languages. Our work extends these findings to the domain of syntactic agreement, revealing that language-agnostic (between English and Spanish) subject number is represented in a one-dimensional subspace. Moreover, we demonstrate that the circuits responsible for solving the number agreement are extremely similar in both languages.

\section{Conclusion}
In this work, we study how Gemma 2B solves the subject-verb agreement task in two different languages, English and Spanish. Through activation patching and direct logit attribution we find that both languages rely on circuits that are highly consistent. Moreover, we provide evidence of an attention head writing a `subject number' signal as a direction from which downstream neurons read to promote the correct verb number continuation. Then, we show this direction has a causal effect, being able to flip the predicted verb number across languages. Finally, we present evidence of similar behavior in other models within the Gemma 1 and Gemma 2 families.

\section{Limitations}
We recognize two main limitations in our study. First, we focused solely on Gemma models. This choice is motivated by its large vocabulary size, which aids in studying multilingual settings. Results obtained on these models do not guarantee they generalize to other model families. Second, our study is limited to two languages: English and Spanish. Although we identified a language-agnostic subject number direction in the model's representation space, demonstrating its generality across these two languages, we cannot conclude that the same applies to all other languages, particularly those that are more linguistically distant.

\section*{Acknowledgments}
The authors acknowledge the anonymous reviewers for the for their valuable insights and constructive feedback. We also thank Neel Nanda for the positive feedback and encouragement to convert this project into a publication. Javier Ferrando is supported by the Spanish Ministerio de Ciencia e Innovación through the project PID2019-107579RB-I00 / AEI / 10.13039/501100011033.

\bibliography{anthology,custom}
\appendix
\newpage
\section{Logit Differences Clean and Corrupted prompts}\label{appx:logit_diffs}

        \begin{figure}[!h]
	\begin{centering}\includegraphics[width=0.45\textwidth]{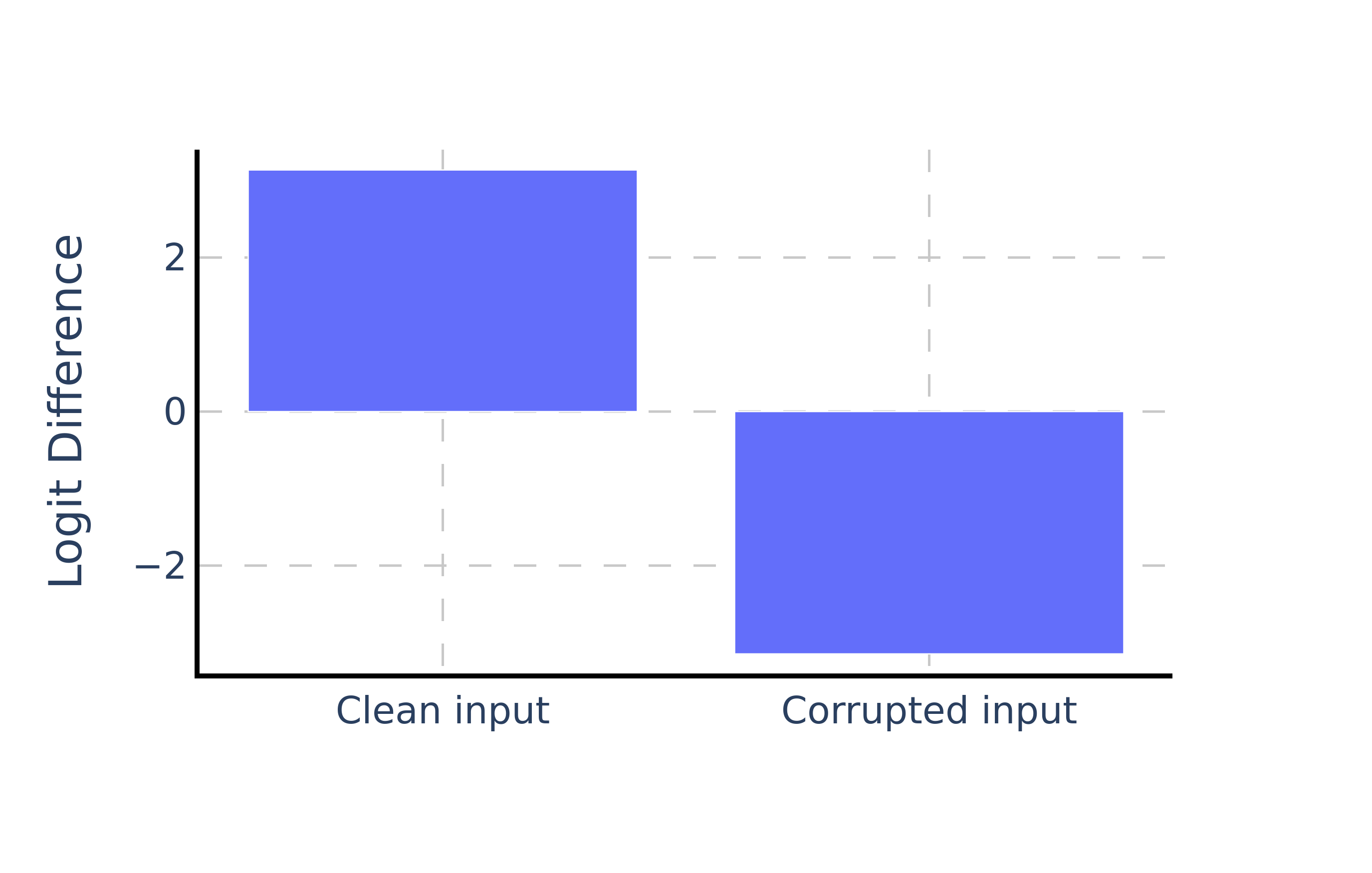}
	\caption{Logit Difference on clean and corrupted inputs. English dataset.}
	\label{fig:english_logit_diffs}
	\end{centering}
\end{figure}

\begin{figure}[!h]
	\begin{centering}\includegraphics[width=0.45\textwidth]{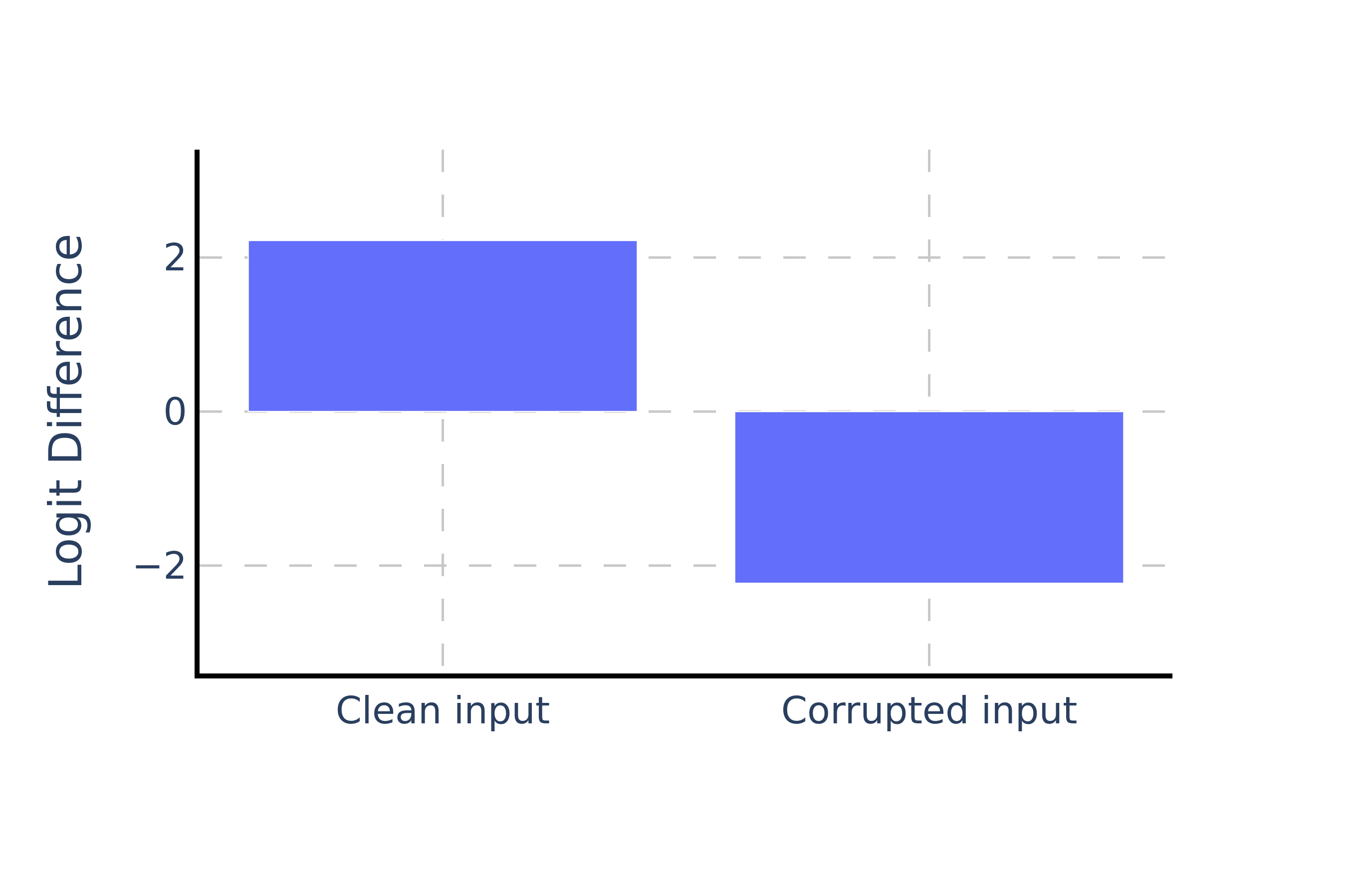}
	\caption{Logit Difference on clean and corrupted inputs. Spanish dataset.}
	\label{fig:spanish_logit_diffs}
	\end{centering}
\end{figure}

\section{Logit Difference by Neurons in MLPs}\label{apx:logit_diff_neurons}
\begin{figure}[!h]
	\begin{centering}\includegraphics[width=0.5\textwidth]{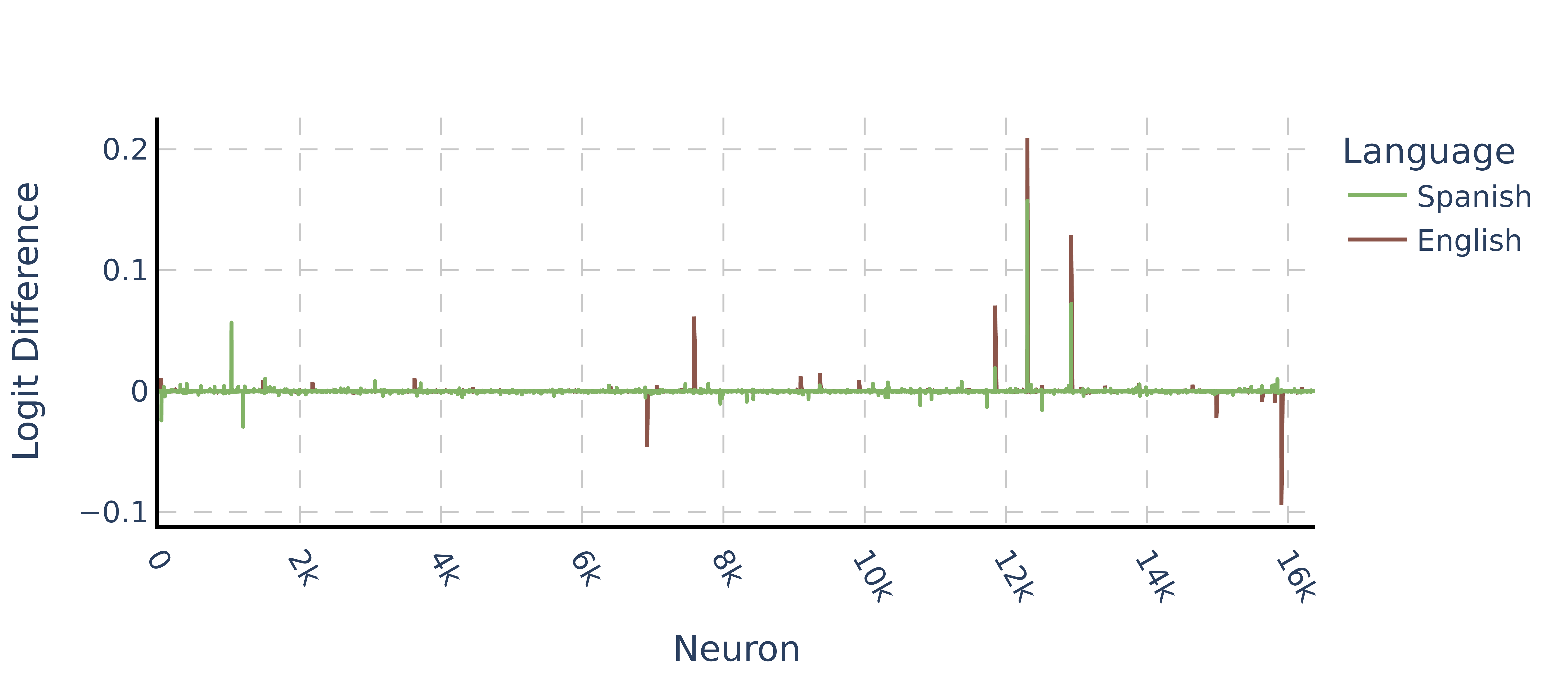}
	\caption{Average contribution to the logit difference by each neuron in MLP15.}
	\label{fig:MLP15_both_neurons}
	\end{centering}
\end{figure}

\begin{figure}[!h]
	\begin{centering}\includegraphics[width=0.5\textwidth]{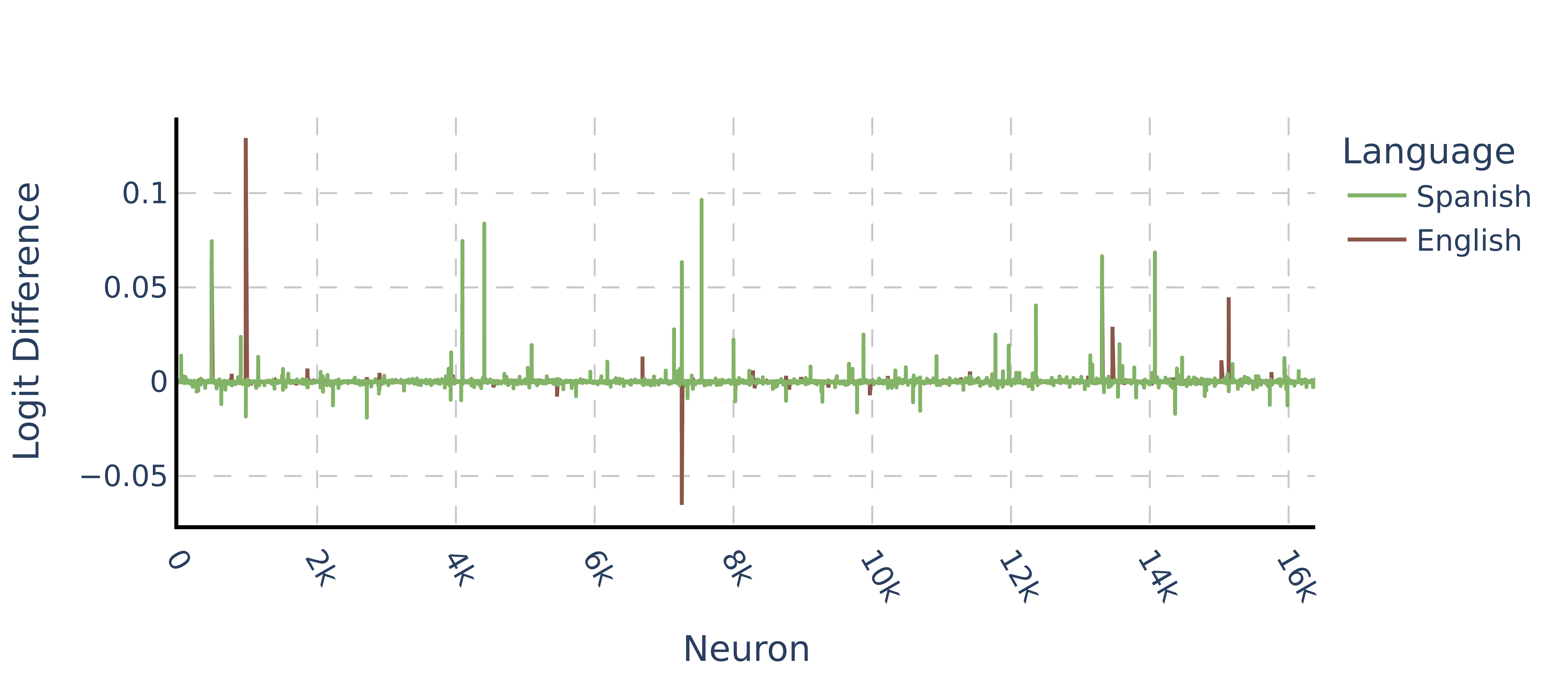}
	\caption{Average contribution to the logit difference by each neuron in MLP16.}
	\label{fig:MLP16_both_neurons}
	\end{centering}
\end{figure}

\begin{figure}[!h]
	\begin{centering}\includegraphics[width=0.5\textwidth]{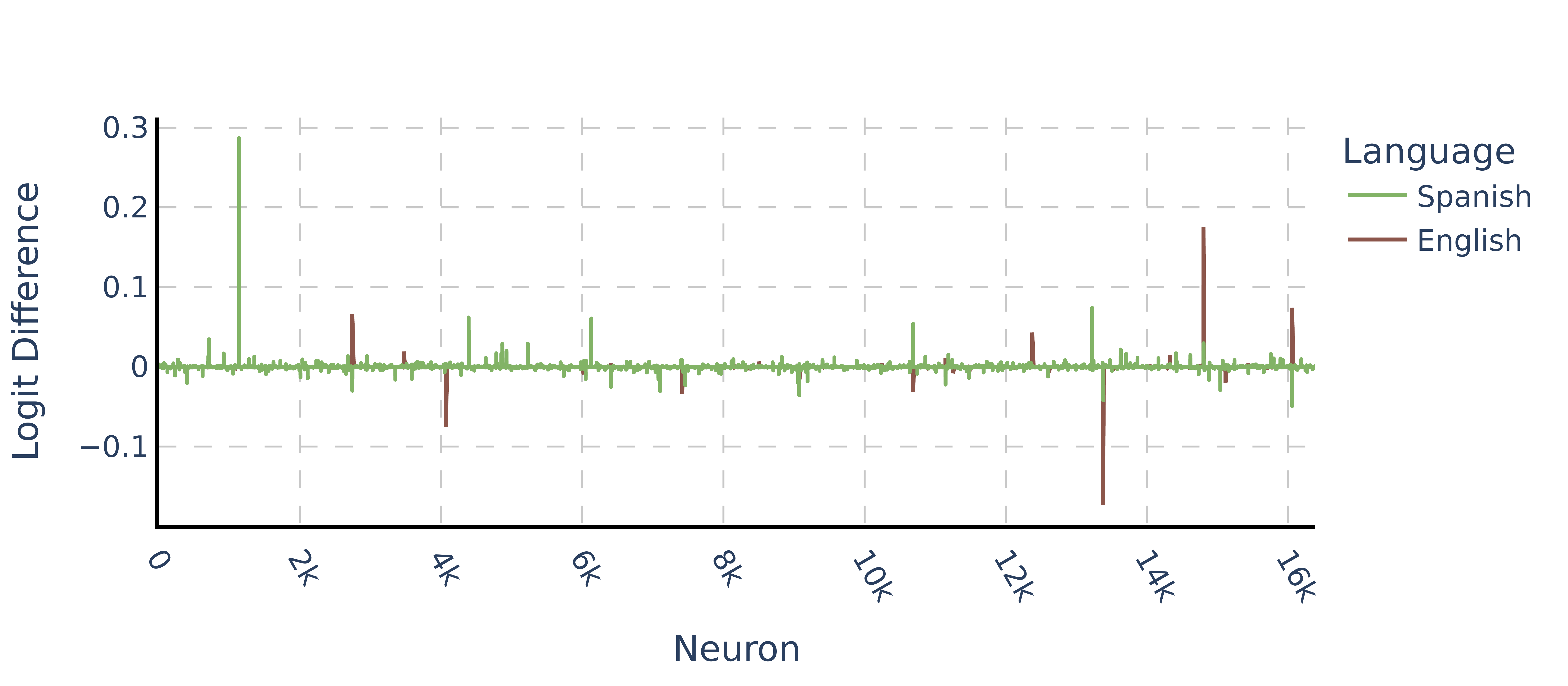}
	\caption{Average contribution to the logit difference by each neuron in MLP17.}
	\label{fig:MLP17_both_neurons}
	\end{centering}
\end{figure}

\newpage
\section{Attention Head L13H7 Composition with Downstream Neurons}\label{sec:dot_products_L13H7}

\begin{figure}[!h]
	\begin{centering}\includegraphics[width=0.43\textwidth]{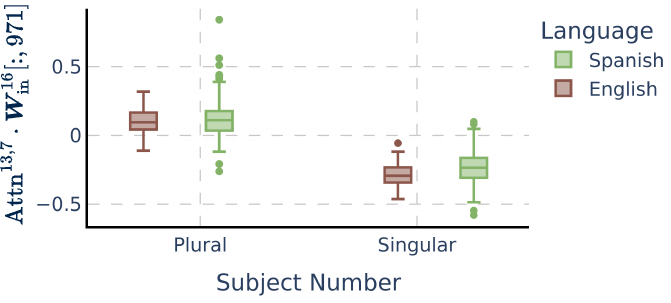}
	\caption{Values of the dot product between the output of attention head L13H7 and the input weights of neuron 971 in MLP16.}
 \label{fig:both_MLP16_neuron971_act_subj_num}
	\end{centering}
\end{figure}

\begin{figure}[!h]
	\begin{centering}\includegraphics[width=0.43\textwidth]{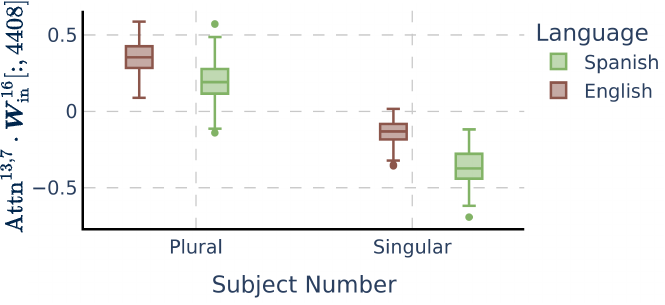}
	\caption{Values of the dot product between the output of attention head L13H7 and the input weights of neuron 4408 in MLP16.}
 \label{fig:both_MLP16_neuron4408_act_subj_num}
	\end{centering}
\end{figure}

\begin{figure}[!h]
	\begin{centering}\includegraphics[width=0.43\textwidth]{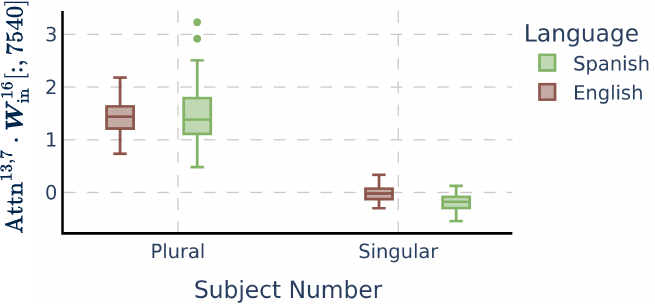}
	\caption{Values of the dot product between the output of attention head L13H7 and the input weights of neuron 7540 in MLP16.}
 \label{fig:both_MLP16_neuron7540_act_subj_num}
	\end{centering}
\end{figure}

\newpage
\section{Neuron 1138 in MLP17}\label{appx:neuron_1138_mlp17}
\begin{figure}[!h]
	\begin{centering}\includegraphics[width=0.43\textwidth]{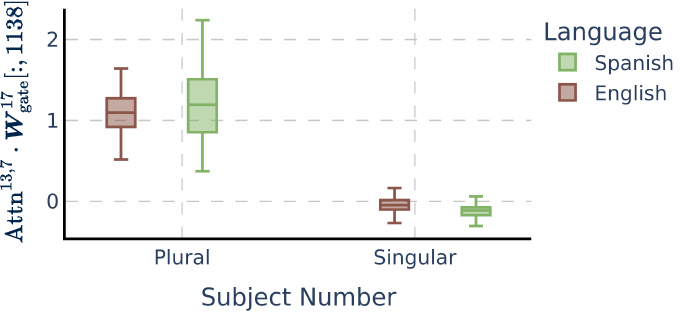}
	\caption{Values of the dot product between the output of attention head L13H7 and the input weights $\mW_{\text{gate}}$ of neuron 1138 in MLP17.}
 \label{fig:both_MLP17_gate_neuron1138_act_subj_num}
	\end{centering}
\end{figure}

Neuron 1138 in MLP17 only activates on sentences with plural subjects. This can be seen in~\Cref{fig:both_MLP17_gate_neuron1138_act_subj_num}, the dot-product of $\mW_{\text{gate}}[:,1138]$ with L13H7 output is negative for singular subjects, meaning that it doesn't activate. In contrast, on plural subjects the dot product of $\mW_{\text{gate}}[:,1138]$ and L13H7 output is positive, and $\mW_{\text{in}}[:,1138]$ is negative, meaning that the neurons fires negatively. In~\Cref{table:promoted_tokens_n1138} we see that the promoted tokens in this case are plural verb forms of multiple non-English languages.
\begin{figure}[!h]
	\begin{centering}\includegraphics[width=0.43\textwidth]{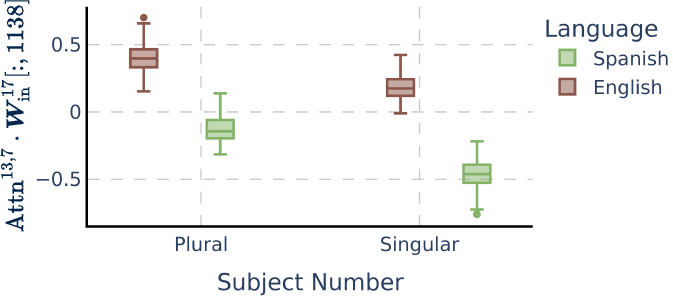}
	\caption{Values of the dot product between the output of attention head L13H7 and the input weights $\mW_{\text{in}}$ of neuron 1138 in MLP17.}
 \label{fig:both_MLP17_in_neuron1138_act_subj_num}
	\end{centering}
\end{figure}

\begin{table}[!h]
    \centering
   \resizebox{0.4\textwidth}{!}{%
    \begin{tabular}{l} \toprule

\multicolumn{1}{l}{\textbf{Top Promoted Tokens \textcolor{red_plotly}{\textit{Negative}} Neuron Activation}}\\ 
 \textbf{`abbiano'},  \textbf{` avevano'},  \textbf{` sembrano'},  \textbf{` avrebbero'},  \textbf{` continúan'},\\  \textbf{` fossero'},  \textbf{` possano'}, \textbf{` poseen'},  \textbf{` tenham'},  \textbf{` terão'}\\
 \textbf{` ont'},  \textbf{` constituyen'},  \textbf{` lograron'}\\
\bottomrule
    \end{tabular}}
    \caption{Top promoted tokens by neuron 1138 in MLP17 based on negative neuron activations.}
    \label{table:promoted_tokens_n1138}
\end{table}
\newpage
\section{MLP and Gated MLP (GMLP)}\label{apx:gmlp}
\begin{figure}[!h]
	\begin{centering}\includegraphics[width=0.45\textwidth]{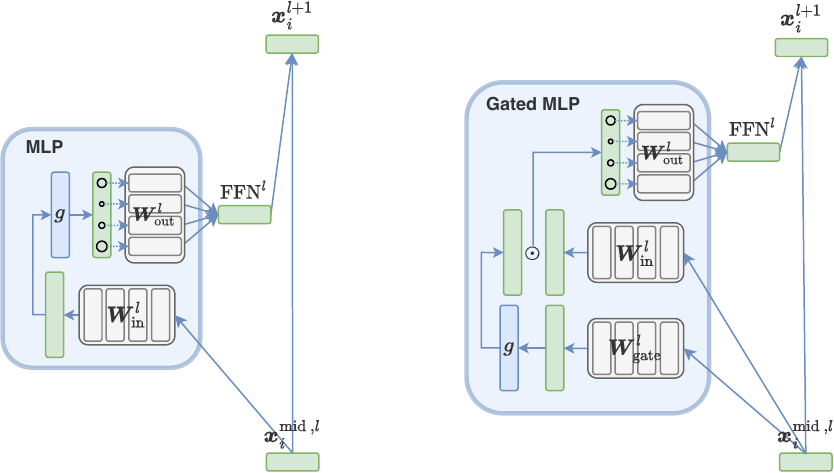}
	\caption{A comparison between the operations performed by the standard MLP and the Gated MLP (GMLP) found in Gemma models.}
	\label{fig:mlp_gmlp}
	\end{centering}
\end{figure}


\section{Attention Patterns Main Heads}\label{appx:attention_patterns}
\begin{figure}[!h]
	\begin{centering}\includegraphics[width=0.5\textwidth]{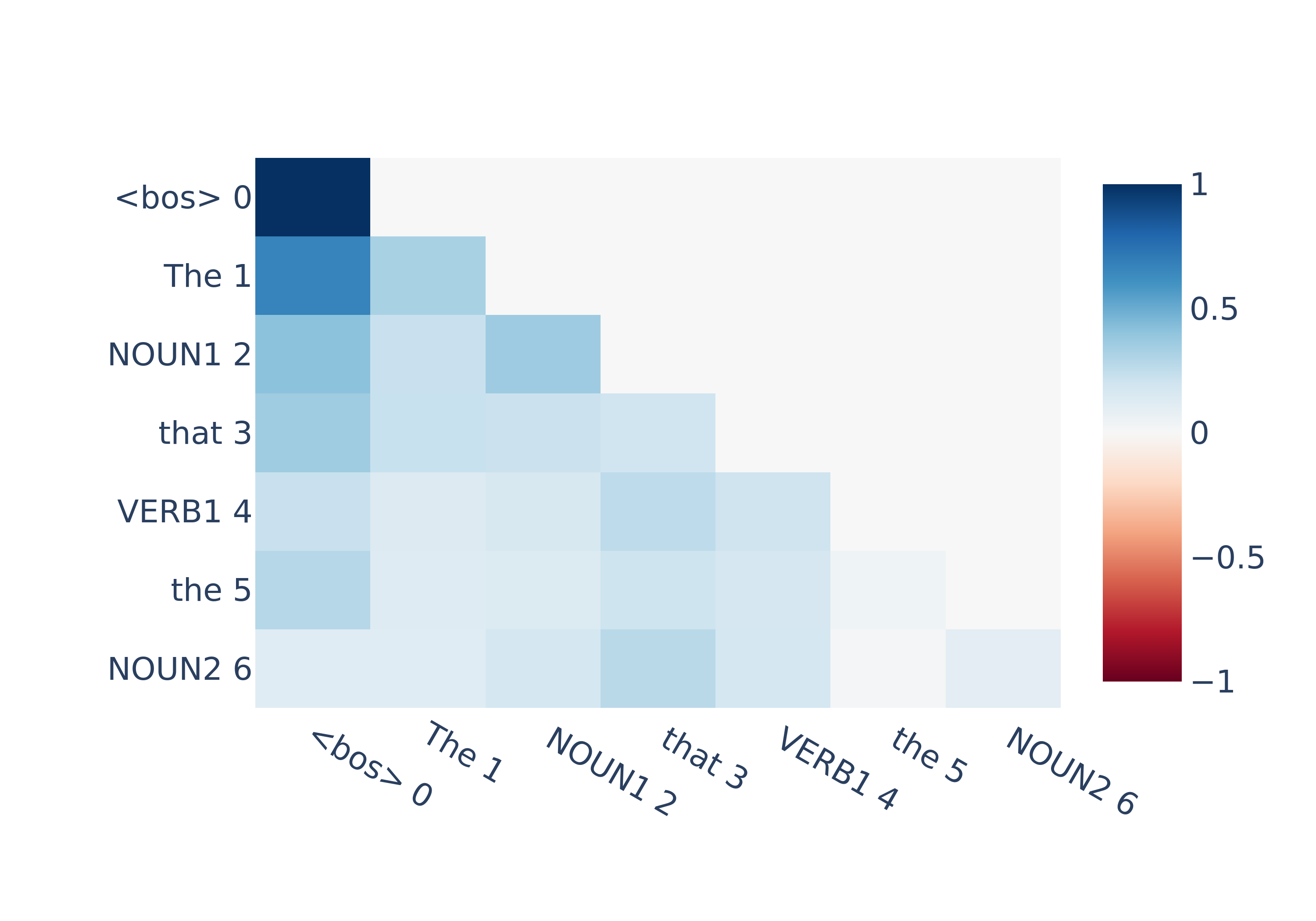}
	\caption{L13H7 average attention patterns (output-value weighted) across the English dataset.}
 \label{fig:L13H7_english_output_value_weighted}
	\end{centering}
\end{figure}

\begin{figure}[!h]
	\begin{centering}\includegraphics[width=0.5\textwidth]{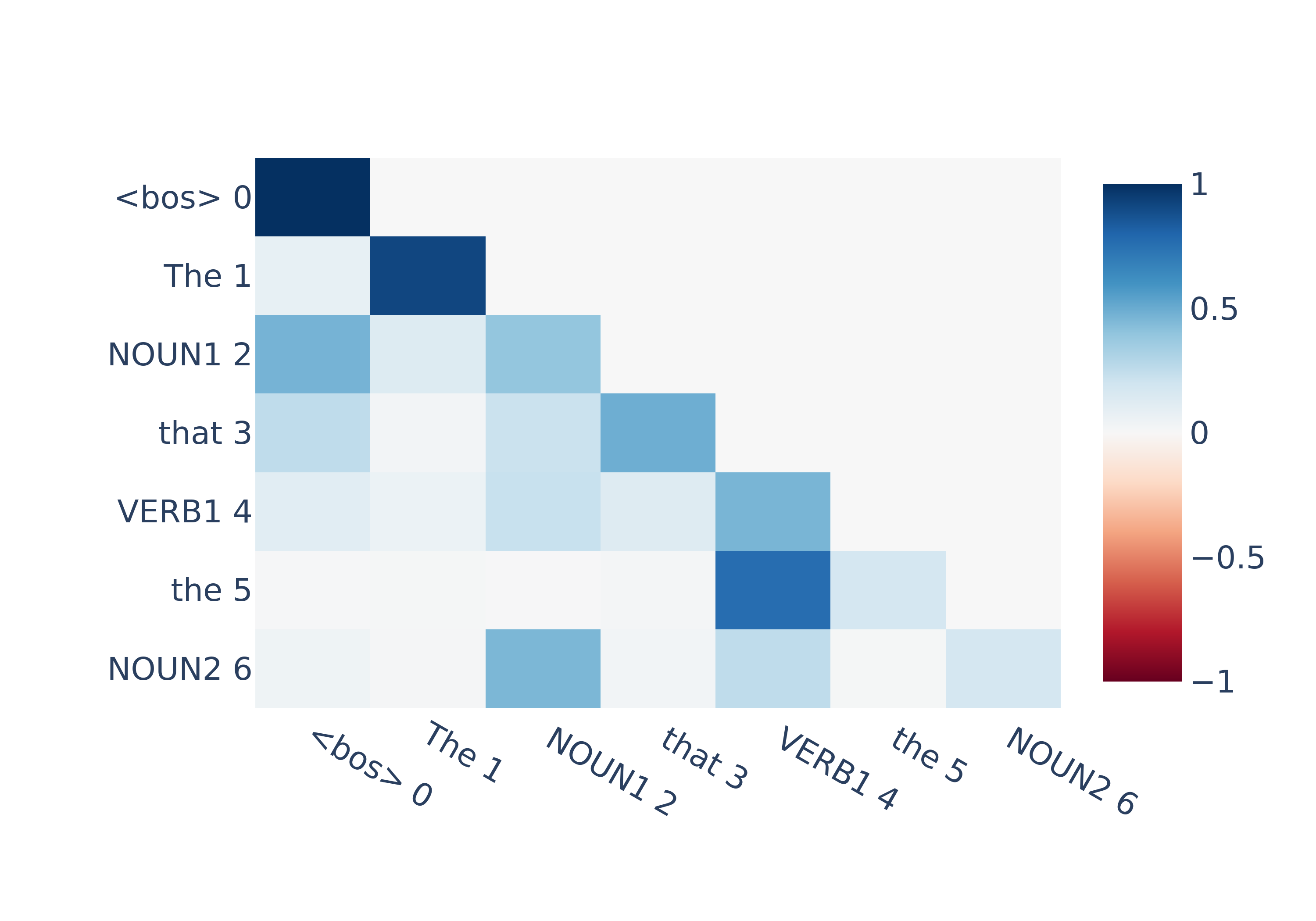}
	\caption{L17H4 Average attention patterns (output-value weighted) across the English dataset.}
 \label{fig:L17H4_english_output_value_weighted}
	\end{centering}
\end{figure}
\vspace{5cm}
\begin{figure}[!h]
	\begin{centering}\includegraphics[width=0.5\textwidth]{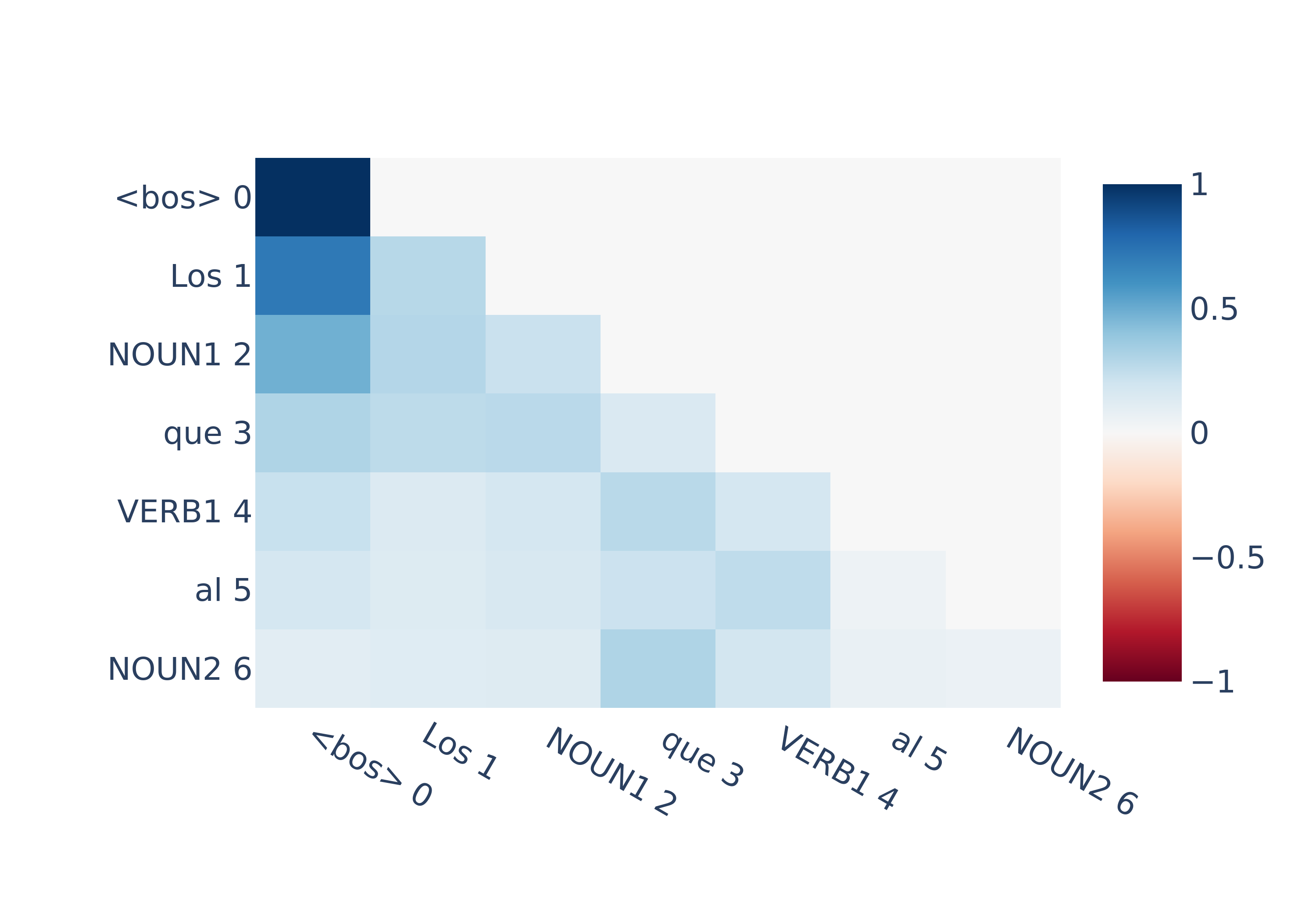}
	\caption{L13H7 average attention patterns (output-value weighted) across the Spanish dataset.}
 \label{fig:L13H7_spanish_output_value_weighted}
	\end{centering}
\end{figure}
\vspace{5cm}
\begin{figure}[!h]
	\begin{centering}\includegraphics[width=0.5\textwidth]{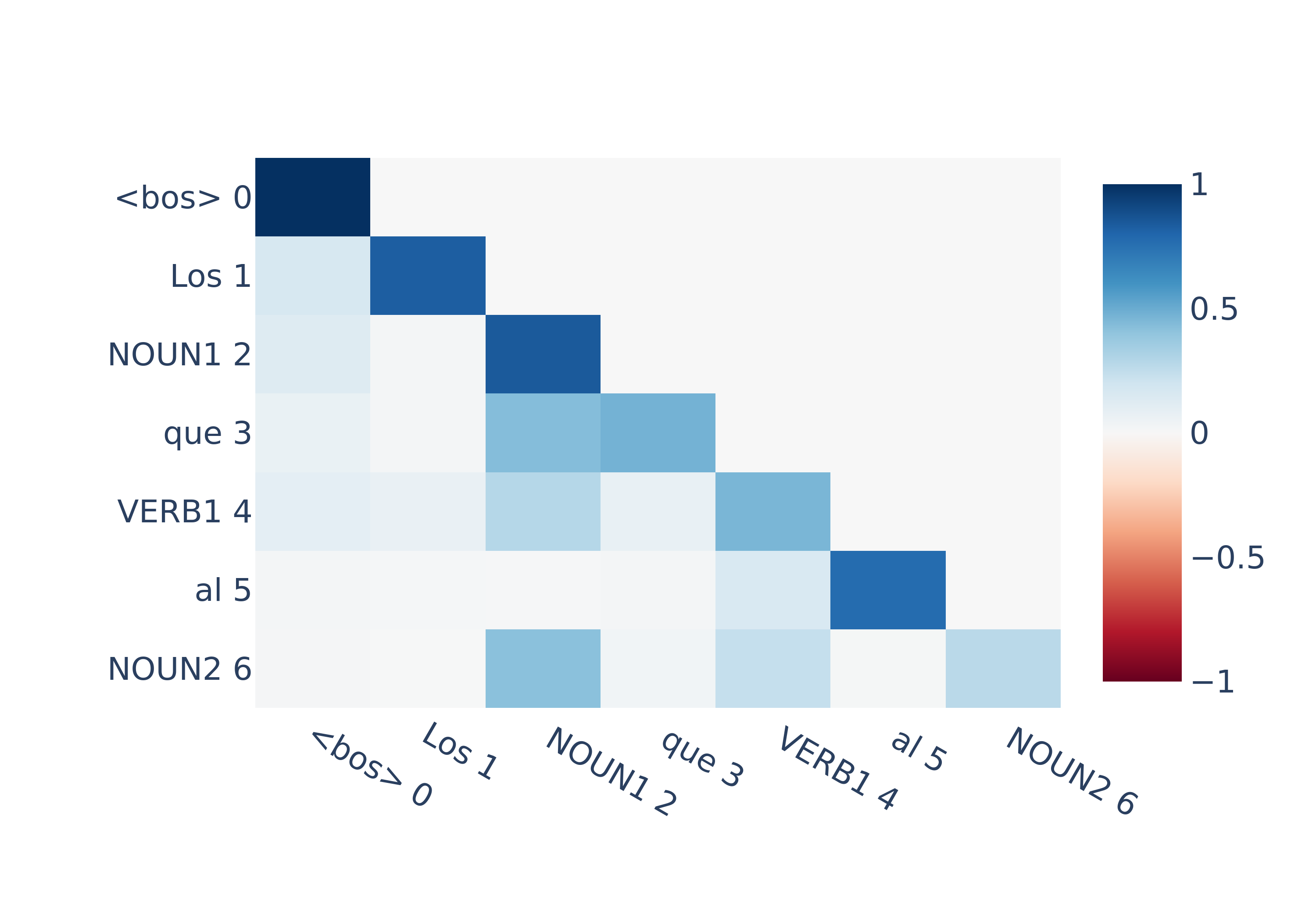}
	\caption{L17H4 Average attention patterns (output-value weighted) across the English dataset.}
 \label{fig:L17H4_spanish_output_value_weighted}
	\end{centering}
\end{figure}

\vfill\eject

\section{Example Top Predicted Tokens in Steering Experiment}\label{apx:predicted_tokens_steering}
\begin{table}[h]
    \centering
   \resizebox{0.4\textwidth}{!}{%
    \begin{tabular}{l} \toprule
    \multicolumn{1}{l}{\textbf{Top 10 Predicted Tokens Before Steering}}\\ 
' se', ' de', ' en', \textbf{' era'}, ' y', ' del', ' ', ',', \textbf{' es'}, \textbf{' fue'}\\
\midrule
\multicolumn{1}{l}{\textbf{Top 10 Predicted Tokens After Steering}}\\ 
' de', ' se', ' en', ' y', ' ', ' del', \textbf{' son'}, ',', ' no', \textbf{' eran'}\\
\bottomrule
    \end{tabular}}
    \caption{Top 10 Predicted Tokens before and after steering a spanish example. In bold are shown spanish forms of the verb `to be'.}
    \label{table:predicted_tokens_steering}
\end{table}

\clearpage
\onecolumn
\section{Activation Patching Results in Gemma 7B}\label{apx:results:gemma7b}

\begin{figure}[h]
\begin{centering}\includegraphics[width=0.98\textwidth]{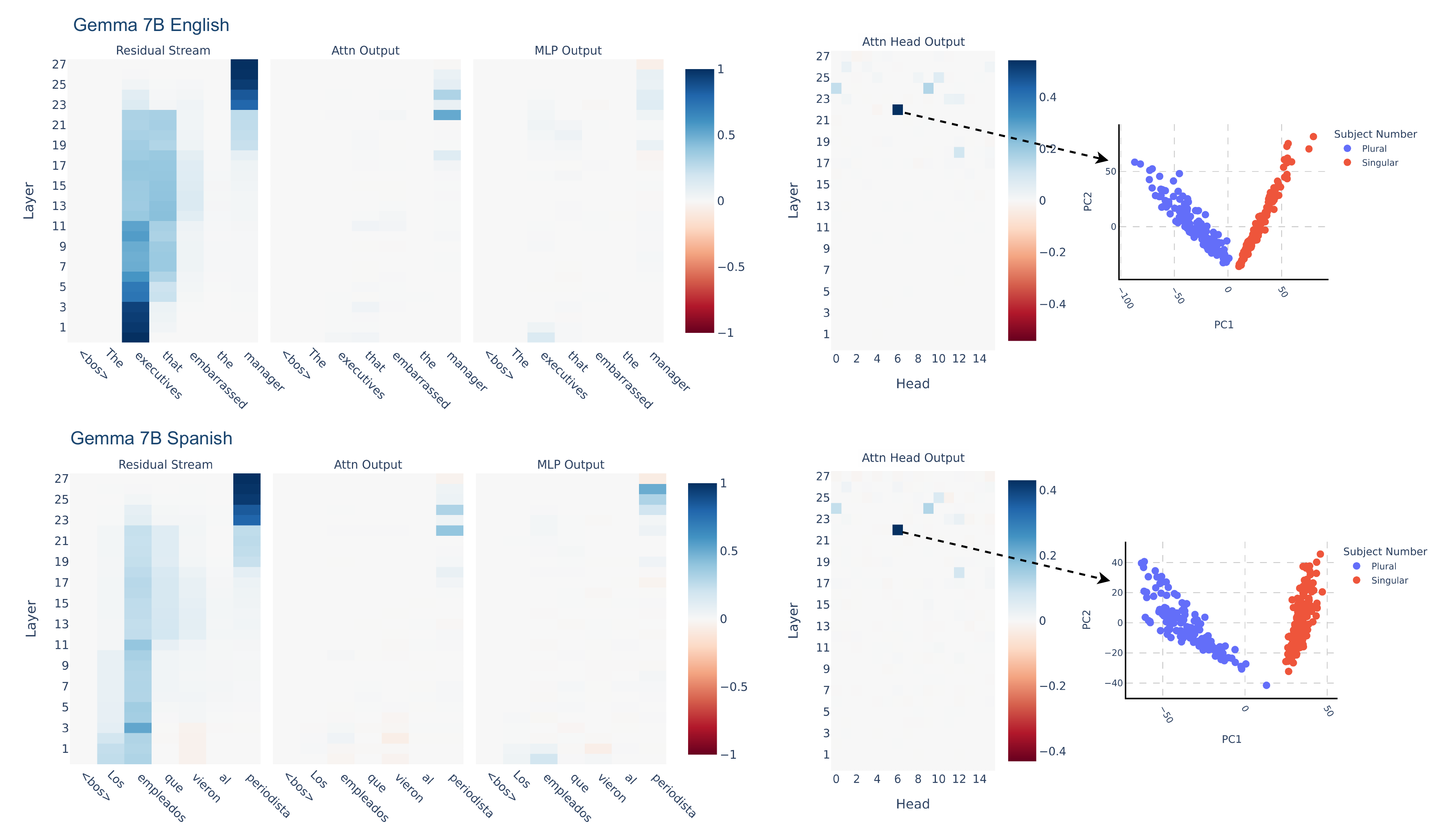}
	\caption{Gemma 7B English (top) and Spanish (bottom) activation patching results. On the right it is shown the outputs of the most relevant attention head (L22H6) in PCA basis. We observe similar patterns to those in Gemma 2B, having a relevant attention head writing the subject number signal into a single direction (PC1).}
	\label{fig:gemma_7b_act_patching_pca}
	\end{centering}
\end{figure}

\end{document}